\documentclass[11pt, logo, onecolumn, copyright, colorlinks=true, allcolors=blue]{nvidiatechreport}
\usepackage{graphicx} % Required for inserting images
\usepackage{subcaption,ragged2e}
\usepackage[round]{natbib}
\usepackage{soul}
\usepackage[normalem]{ulem}
\usepackage[utf8]{inputenc} % allow utf-8 input
\usepackage[T1]{fontenc}    % use 8-bit T1 fonts
\usepackage{url}
\usepackage{tikz}
\usepackage{enumitem}
\usetikzlibrary{positioning}
\usepackage{array}
\usepackage{booktabs}
\usepackage{wrapfig}
\usepackage{caption} % For a better control over table captions
\usepackage{listings}
\usepackage{adjustbox}
\usepackage{footnote}
\usepackage{multirow}
\usepackage{amsmath}
\usepackage{amsfonts}
\usepackage{breakcites}
\usepackage{blindtext}
\usepackage{svg}
\usepackage[most]{tcolorbox}
\usepackage{tabularx}
\usepackage{makecell}
\usepackage{graphicx}
\newcommand{\sred}[1]{\textcolor{red}{\texttt{#1}}}
\newcommand{\sblue}[1]{\textcolor{blue}{\texttt{#1}}}
\newcommand{\sgreen}[1]{\textcolor{ForestGreen}{\texttt{#1}}}
\newcommand{\spurple}[1]{\textcolor{purple}{\texttt{#1}}}
\title{NVIDIA Nemotron-Parse 1.1}
\author{\large NVIDIA}
\date{}

\begin{document}

\begin{abstract}
\large \textbf{Abstract.}
We introduce Nemotron-Parse-1.1, a lightweight document parsing and OCR model that advances the capabilities of its predecessor, Nemoretriever-Parse-1.0 \citep{karmanov2025eclairextractingcontent}. Nemotron-Parse-1.1 delivers improved capabilities across general OCR, markdown formatting, structured table parsing, and text extraction from pictures, charts, and diagrams. It also supports a longer output sequence length for visually dense documents. As with its predecessor, it extracts bounding boxes of text segments, as well as corresponding semantic classes.

Nemotron-Parse-1.1 follows an encoder-decoder architecture with 885M parameters, including a compact 256M-parameter language decoder. It achieves competitive accuracy on public benchmarks making it a strong lightweight OCR solution. We release the model weights publicly on \href{https://huggingface.co/nvidia/NVIDIA-Nemotron-Parse-v1.1}{Huggingface}, as well as an optimized \href{https://build.nvidia.com/nvidia/nemotron-parse}{NIM container}, along with a subset of the training data as part of the broader \href{https://huggingface.co/datasets/nvidia/Nemotron-VLM-Dataset-v2}{Nemotron-VLM-v2} dataset. Additionally, we release \href{https://huggingface.co/nvidia/NVIDIA-Nemotron-Parse-v1.1-TC}{Nemotron-Parse-1.1-TC} which operates on a reduced vision token length, offering a 20\% speed improvement with minimal quality degradation.

\end{abstract}

\maketitle

\section{Introduction}
In the recent years, document-level OCR has developed beyond simple extraction of plain text characters from an image. Modern applications such as Large Language Models, retrieval systems, question-answering solutions, demand a richer representation, which includes layout, reading order, semantic classes (such as captions or footnotes), formulas, tables, and understanding of multi-column/multi-page structure. 

A number of models were introduced in recent years, ranging from pipeline-based solutions to end-to-end Vision-Language models \citep{feng2025dolphin, marker, mathpix, wang2409mineru, li2025monkeyocr, cui2025paddleocr, Nougat, nassar2022tableformer, nassar2025smoldocling, poznanski2025olmocr, GOT, ocrflux, dotsocr}.

Pipeline solutions often rely on brittle multi-stage pipelines with each stage responsible for a subtask of the overall document extraction process, hence achieving versatility at the cost of the lower throughput. At the same time, end-to-end models benefit from fast inference speeds while often not performing equally well on all subtasks associated with document extraction simultaneously, i.e., general OCR, structured text formatting, mathematical equations, extraction of complex tables, prediction of bounding boxes and semantic classes of text blocks.

To close this gap, we introduce Nemotron Parse 1.1, a successor to Nemoretriever-Parse 1.0 and Eclair~\citep{karmanov2025eclairextractingcontent}, which is an end-to-end vision-language model capable of extracting formatted text (Markdown/LaTeX), bounding boxes of text blocks, and semantic classes for each block while preserving the reading order.  
Beyond the full-capability Nemotron Parse 1.1 model, we also introduce Nemotron Parse 1.1 with token compression (TC), a streamlined variant designed for applications that prioritize speed without majorly sacrificing output quality. Nemotron-Parse-TC maintains nearly the same level of accuracy as the full model, while offering reduced latency and lower computational overhead. This makes it particularly well-suited for large-scale batch processing, edge deployments, or interactive systems where rapid response times are critical. Despite its lightweight architecture, Nemotron-Parse-TC preserves support for core document understanding features—including layout-aware text extraction, semantic block classification, and consistent reading-order generation—providing an efficient alternative for production environments that demand both performance and precision. Additionally, Nemotron-Parse-TC offers an improved reading order, with all page elements, including floating ones, following the page ordering.

\section{Nemotron-Parse}

\subsection{Model Architecture}

Nemotron-parse follows an encoder-decoder transformer architecture. The \textbf{vision encoder}, denoted as $\mathcal{E}$, is initialized from RADIO~\citep{radio, Heinrich_2025_CVPR} which follows a ViT-H /16 \citep{vit} architecture (657M parameters), and maps an image  $\mathbf{I} \in \mathbb{R}^{3 \times H \times W}$ to a latent representation $\mathbf{Z} \in \mathbb{R}^{N \times d}$, where $d$ is the hidden dimension and $N$ is the sequence length. 

The vision \textbf{neck} $\mathcal{N}$ consisting of horizontal convolutional kernels of size $1\times 4$ and stride $1\times 4$ then reduces the dimensionality of the latent space as well as the sequence length. For an input image of $1648\times2048$ this reduces the sequence length to $3200$. We additionally concatenate the summary token of RADIO to the sequence.

For Nemotron-Parse-TC, we additionally apply pixel-shuffle on top of the compressed sequence, further reducing the sequence length to $833$ tokens, hence achieving a total of $\times16$ reduction.

% \hl{TODO: add about token reduction. Maybe also include ablation of 1x4 vs 2x2 vs full pixshuffle since we have it for stage 1 anyways}
% 1x4: 0.89 / 0.186 / 0.67. 2x2: 0.88 / 0.1939 / 0.66

The \textbf{decoder}, denoted as $\mathcal{D}$, uses mBART \citep{mbart} architecture reduced to 10 layers and with tied wights, and predicts text tokens $\mathbf{T} = \{t_{P+1}, t_{P+2}, \dots, t_L\}$ by conditioning on the latent encoder representation, $\mathcal{N}(\mathbf{Z})$, and the context $t_{<i}$, $P(t_i | \mathcal{N}(\mathbf{Z}), t_{<i})$, where $\mathbf{Z} = \mathcal{E}(\mathbf{I})$ and $\{t_1, t_2, \dots, t_P\}$ are the prompt tokens and where $L$ is the prompt-augmented sequence length. The model has 
885M parameters in total. %884766720

\subsubsection{Positional embeddings}
To enable large-context inference, we train and evaluate the model without positional embeddings in the decoder. We find that the network achieves comparable accuracy to models trained with positional embeddings, while allowing inference with significantly longer context lengths.

In the design of Vision-Language Models (VLMs) for OCR, our decision to omit positional embeddings in the LLM decoder is motivated by the ability of decoder-only transformer architectures to implicitly encode position. While transformers are often described as permutation-invariant and thus dependent on positional embeddings, in causal decoder-only models the attention mask already provides positional cues: each token can only attend to preceding elements, which enables the model to infer its location in the sequence \citep{kazemnejad2023impact, zuo2025position}.

This simplifies the architecture by removing additional positional parameters and aligns with the multimodal nature of OCR, where visual tokens from the image encoder carry spatial structure relative to the document layout. Without an extra 1D positional signal, the decoder avoids possible interference between sequence-based embeddings and the 2D spatial information already present in the visual features.

For OCR, this reduced reliance on explicit positional encoding can improve generalization and computational efficiency across different document lengths, from short receipts to long academic pages with complex tables. No Positional Encoding (NoPE) approaches have been reported to generalize better to longer sequences, since they avoid interpolation issues that arise with explicit embeddings \citep{kazemnejad2023impact, zuo2025position}.
% From Ilia

\subsubsection{Multi-token inference}

Autoregressive models, including those used for OCR, operate by decoding one token at a time, leading to slow inference—especially for text-dense images. To address this, we repurpose our solution from Nemotron-Parse for multi-token generation, predicting $n$ tokens simultaneously \citep{gloeckle2024better}.

During training, for predicting $m$ tokens we add $m-1 \times 2$ additional linear layers. We adopt a simple architecture, where given the context of size $n$, the logits for $n+1^{st}$ token are obtained following standard architecture from $\mathbf{h}_n$, i.e., the final hidden state of the $n^{th}$ token, and for subsequent $n+2..m$ tokens is obtained as $\mathit{l_{head}}(\mathit{l_1}(\mathbf{h_n} + \mathit{l_2}(\mathbf{e_{n+1}})))$, where $\mathbf{e_{n+1}}$ is the embedding of $n+1^th$ token predicted by the preceding head, $\mathit{l_1}$ and $\mathit{l_2}$ are Linear layers, and $\mathit{l_{head}}$ refers to the decoder head. During training, we use teacher forcing for token embeddings of additional $n+2..m$ tokens. At inference, decoding proceeds greedily without token verification.

We find that adoption of the multi-token training strategy additionally allows to achieve improved accuracy in the default single-token inference setup, compared to the models trained with a standard protocol.

\subsection{Prompts and Output Format}

\subsubsection{Input prompts}

Similarly to Eclair \citep{karmanov2025eclairextractingcontent}, we train the model jointly on heterogeneous datasets that provide different supervision signals (plain or formatted text, bounding boxes, and semantic classes). To unify these sources, we use a fixed prompt interface and assign, for each training sample, the prompt that exactly matches the annotations available in its dataset. This lets the model learn a single conditional interface while leveraging all datasets efficiently.

At the core are three independent prompt tokens that define the requested outputs, yielding the eight valid combinations used in training and inference:
\begin{itemize}
\item Text formatting prompts
\begin{itemize}
\item \verb|<output_markdown>|: text is formatted as Markdown, and formulas and Tables are formatted as LaTeX. Inline formulas that do not require any LaTeX syntax so be represented (e.g., consisting only of characters and subscripts/superscripts) remain in markdown format for versatility.
\item \verb|<output_plain>|: emit unformatted text; inline formulas are plain text.
\item \verb|<output_no_text>|: output no text.
\end{itemize}
\item Bounding Box prompts
\begin{itemize}
\item \verb|<predict_bbox>|: return bounding boxes for detected elements.
\item \verb|<no_bbox>|: output no bounding boxes.
\end{itemize}
\item{Class prompts}
\begin{itemize}
\item \verb|<predict_classes>|: return semantic class labels for each box. This option is used only together with \verb|<bbox>|.
\item \verb|<no_classes>|: suppress class labels.
\end{itemize}
\end{itemize}

During training, we map each dataset’s label set to a compatible prompt, ensuring only valid combinations are used (we exclude the trivial “no output” case and any request for classes without boxes). During inference, we define the maximal-information prompt (MIP) as: \begin{lstlisting}[basicstyle=\ttfamily,breaklines=true]
<output_markdown><predict_bbox><predict_classes>,
\end{lstlisting}
i.e., prediction of formatted text, bounding boxes, and semantic classes.

\subsubsection{Output format}

Nemotron-Parse predicts bounding boxes of the semantic blocks in the form of relative coordinates, in a scale of $1024\times1280$. These bounding boxes are predicted in a canonical reading order, that includes Page-Header elements at the start of the page, followed by Text, Section-Header, List-Item, Title and Formula elements in the order as they would be read by a person looking at the given page, followed by Footnotes, Page-Footers, Tables, Pictures, and Captions in the end. Nemotron-Parse-TC improves upon this canonical reading order, and also includes non-reading-order (floating) elements, i.e., Footnotes, Page-Footers, Tables, Pictures, and Captions within the natural ordering of the page.

In the maximal information setting the output is in the following format:

\begin{small}
\begin{center}
\sred{<x\_(\string\d+)>}%
\sred{<y\_(\string\d+)>}%
\sgreen{(.*?)}%
\sblue{<x\_(\string\d+)>}%
\sblue{<y\_(\string\d+)>}\newline
\spurple{<class\_([\string^>]+)>}
\end{center}
\end{small}

\noindent where the the \sred{first group} denotes the coordinates of the top-left corner, the \sgreen{second group} denotes the text contained within the bounding box, the \sblue{third group} denotes the coordinates of the bottom-right corner, and the \spurple{final group} represents the semantic class. For example, \sred{<x\_0.1152><y\_0.2586>}%
\sgreen{\# NVIDIA Nemotron-Parse 1.1}%
\sblue{<x\_0.8799><y\_0.2797>}
\spurple{<class\_Title>}.

\section{Training and Data}

\subsection{Datasets}

Nemotron-Parse is trained on a combination of synthetic, public, and human-annotated data. The overview of the data used for training is provided in Table~\ref{tab:training-datasets} and we provide brief descriptions of internally curated and synthetic data below.

\begin{table}[h]
    \centering
    \small
    %\footnotesize
    \begin{tabular}{lccc}
        \toprule
        \textbf{Dataset/source}         & \textbf{Size}   & \textbf{Modality} & \textbf{Languages}\\
        \midrule
        \textbf{Multilingual arXiv}        & 8.3M              & {Structured, Boxes, Classes} & \makecell{English, Chinese, German\\ Spanish, French, Italian\\ Japanese} \\\hline
        \makecell{\textbf{SynthTabNet}\\\citep{nassar2022tableformer}}     & 480K            & \makecell{Structured, Boxes, Classes} & English \\\hline
        \makecell{\textbf{DocLayNet}\\~\citep{doclaynet2}}       & 56K             & \makecell{Plain+Structured, Boxes, Classes} & English \\\hline
        \textbf{Common Crawl samples} & 255k & \makecell{Plain+Structured, Boxes, Classes} & Various \\ \hline% includes all human labeled, azure, and adobe 
        \textbf{Synthetic tables } & 26K & \makecell{Structured, Boxes, Classes} & English \\ \hline % synthtables 
        \textbf{\makecell{Multilingual \\Synthetic OCR data}} & 3.5M & \makecell{Plain+Structured, Boxes, Classes} & \makecell{English, Chinese, Japanese, \\ Korean, Latin, Greek} \\ \hline
        \textbf{\makecell{Multilingual \\ Wikipedia OCR data}} & 9.5M & \makecell{Structured, Boxes, Classes} & \makecell{English, French, German, \\ Spanish, Italian, Dutch, \\ Portugese, Japanese, Korean, \\ Chinese} \\ \hline
        \makecell{\textbf{Pubtables} \\ \citep{smock2022pubtables}} & 585K & \makecell{Structured, Boxes, Classes} & English \\ \hline
        \makecell{\textbf{Fintabnet} \\ \citep{fintabnet}} & 91.5K & \makecell{Structured, Boxes, Classes} & English\\ \hline
        \makecell{\textbf{TabRecSet} \\ \citep{yang2023large}} & 38.2K & \makecell{Structured, Boxes, Classes} & English, Chinese  \\ \hline
        %\textbf{RVL (funsd)} & & & \\ \hline % not sure if needs to be separate category
        
        \bottomrule
        %\textbf{Total} & 6.176M & \\

    \end{tabular}
    \vspace{-3mm}
    \caption{Summary of the datasets used to train Nemotron-Parse, including a description of the maximum information available in the annotations of each dataset. \vspace{-12pt}}
    \label{tab:training-datasets}
\end{table}

\paragraph{NVpdftex pipeline}

At the core of the data utilized for training the Nemotron-Parse, we build a data generation pipeline inspired by Nougat~\citep{Nougat}. Unlike their pipeline which builds supervision by turning arXiv papers in {\LaTeX} format into HTML with LaTeXML and then into markdown, our pipeline couples {\LaTeX} compilation with structured-output extraction in a single pass, preserving tight alignment between the rendered page and the text down to character-level bounding boxes and enabling per-box semantic labels.

We extend the open-source {\TeX} Live toolchain which allows us to intercept node and character creation, hbox/vbox allocations, token reads, and page output events, leading to preservation of accurate bounding boxes, semantic classes, and reading order.

Using this integrated method, we produced a high-quality large-scale ground-truth corpus of documents referred to as NVpdftex. Additionally, we open-source the generation pipeline for the community \url{https://github.com/NVIDIA-NeMo/Curator/tree/experimental/experimental/nvpdftex}.

To improve multilingual capabilities of Nemotron-Parse we generate additional data by applying machine translation to NVpdftex dataset in 6 languages. We also employ LaTeX-level augmentations of fonts, color, and layout to increase the diversity of the datasets.

\paragraph{DocLayNet}
DocLayNet \citep{doclaynet2} is a public dataset used for layout analysis. On top of the existing annotations, we additionally autolabel the reading order of the text inside the DocLayNet samples, text inside images, as well as markdown formatting, and formatting of Table objects and Formulas. The resulting blend includes a combination of variants of different output formats, including both plaintext and markdown/LaTeX.

\paragraph{Common Crawl Data}

We utilize a set of diverse data samples from Common Crawl which are annotated in plaintext format along with bounding boxes and semantic class labels by human experts. We further autolabel text inside images and markdown formatting to increase the format diversity of the data. In case of autolabeling, structured formatting labels are derived by treating each individual bounding box crop as an individual image and running inference with stage-1 trained Nemotron-Parse. The low-quality predictions are filtered out and blanked on the image on the basis of edit distance of the formatting-stripped output to the plaintext labels. Page-level formatting, such as multi-level headers, are obtained by running full-page inference and aligning global header formatting with individual text blocks heuristically.

\paragraph{Synthetic tables}
To ensure strong table extraction performance, a significant portion of table data is synthetically generated with various styles and layouts. The data is generated in HTML format, and further converted to LaTeX, and rendered on a page to produce a corresponding image. We capture various layouts, text formatting, sparsity levels, presence of checkboxes, etc. 

\paragraph{Multilingual dense OCR data}

Upon observation that oftentimes models struggle with dense OCR, we synthetically generate dense text of multiple languages and render it on the image. This includes random words, characters, and symbols, in 6 different languages. 

\paragraph{Multilingual Wikipedia OCR data}

An additional source of multilingual data is obtained from Wikipedia text in multiple languages that is converted to LaTeX formatting and augmented with font, background, and color augmentations. 

\paragraph{Pubtables, Fintabnet, TabRecSet}
We additionally utilize publicly available table extraction datasets \citep{smock2022pubtables, fintabnet, yang2023large} where we convert the tables from HTML to LaTeX format and autolabel the remaining elements (if present) in the pages to follow the Nemotron-Parse format. 

A large portion of the synthetic and human-labeled datasets have been released as part of the \href{https://huggingface.co/datasets/nvidia/Nemotron-VLM-Dataset-v2}{Nemotron-VLM-Dataset-V2} release.

\section{Experimental results}

\subsection{OCR and reading order evaluation}
\begin{table*}[t]
    \centering
    %\scriptsize
  \begin{tabular}{l|c|cc}
\toprule
\textbf{Method} & \textbf{\shortstack{Mask\\out}} & \textbf{WER} ↓& \textbf{F1} ↑ \\
\midrule
{\shortstack{\vspace{-2pt}\\Kosmos-2.5 \citep{Kosmos} (ocr-mode)}}& +  & 0.195 & 0.937 \\ 
{\shortstack{\vspace{-2pt}\\Kosmos-2.5 \citep{Kosmos} (md-mode)}}& +  & 0.249 & 0.890 \\
{\shortstack{\vspace{-2pt}\\GOT \citep{GOT} (ocr-mode)}}& +  & 0.302 & 0.818 \\ 
{\shortstack{\vspace{-2pt}\\GOT \citep{GOT} (md-mode)}} & +  & 0.259 & 0.879 \\
\midrule
{\shortstack{\textbf{Nemotron-Parse-MIP}}}& - & 0.109 & \textbf{0.958} \\ 
{\shortstack{\vspace{-2pt}\\\textbf{Nemotron-Parse-MIP}}}& + & \textbf{0.102} & 0.957 \\ 
{\shortstack{\vspace{-2pt}\\\textbf{Nemotron-Parse-TC-MIP}}}& - & 0.111 & 0.953 \\
{\shortstack{\vspace{-2pt}\\\textbf{Nemotron-Parse-TC-MIP}}}& + & 0.121 & 0.949 \\
\bottomrule
\end{tabular}
\vspace{-2mm}
    \caption{Evaluation results on an internal test set.}
    \label{testsetv3}
\end{table*}

We assess reading-order accuracy of Nemotron-Parse on an internally curated, human-labeled set of 789 PDF pages \citep{karmanov2025eclairextractingcontent} drawn from magazines, books, and the Common Crawl corpus \citep{sebastian_spiegler_statistics_2013}. This test set reflects the layout diversity seen in DocLayNet \citep{doclaynet2}, an human annotators followed DocLayNet’s labeling scheme, with a key addition of the explicit reading-order annotations.

Using this benchmark, we compare Nemotron-Parse against baselines Kosmos-2.5 \citep{Kosmos} and GOT \citep{GOT} and report the results in Table~\ref{testsetv3}. We normalize the outputs of all the methods and exclude tables, equations, and {\TeX} commands from evaluation in this benchmark. Additionally, for GOT (md), we also mask-out headers and footers from the images, as their model seems to ignore these elements. Because both baselines offer two output modes—plain OCR and markdown—we evaluate Nemotron-Parse against each of these modalities. 

\begin{table}[]
\small
\begin{tabular}{l|cccc}
\toprule
\textbf{Extractor}               & \textbf{\makecell{OCR/F1 \\ Score $\uparrow$}} & \textbf{\makecell{Text-Only RO/ \\ Edit Dist. $\downarrow$}} & \textbf{\makecell{Text-Only RO/ \\METEOR $\uparrow$}} & \textbf{\makecell{Text-Only RO/ \\ BLEU $\uparrow$}} \\
\midrule
Pdfium                           & 0.0036                           & 0.9993                                         & 0.0083                                   & 0.0000                                 \\
Docling                          & 0.6744                           & 0.4300                                         & 0.6331                                   & 0.4651                                 \\
Gemini Flash 2.0                 & \textbf{0.9915}                  & \textbf{0.0125}                                & \textbf{0.9934}                          & \textbf{0.9828}                        \\
Mistral                          & 0.9729                           & 0.0189                                         & 0.9831                                   & 0.9560                                 \\
LandingAI Document Agent         & 0.9524                           & 0.0699                                         & 0.9428                                   & 0.8945                                 \\
Marker                           & 0.9696                           & 0.0322                                         & 0.9795                                   & 0.9451                                 \\
SmolDocling                      & 0.9588                           & 0.0352                                         & 0.9728                                   & 0.9343                                 \\ 
\midrule
\textbf{Nemotron-Parse-1.1}      & 0.9785                           & 0.014                                          & 0.9858                                   & 0.9623                                 \\
\textbf{Nemotron-Parse-1.1-TC} & 0.9755                           & 0.014                                          & 0.9838                                   & 0.9582                \\
\bottomrule
\end{tabular}
\caption{OCR metrics on GOT~\citep{GOT} benchmark}\label{tab:GOT}
\end{table}

Next, we compare Nemotron-Parse-1.1 to other popular OCR solutions on the GOT benchmark \citep{GOT}. The results are reported in Table~\ref{tab:GOT}. As can be seen, Nemotron-Parse shows a strong performance, being outperformed only by Gemini Flash 2.0.

Further, we evaluate Nemotron-Parse on widely-adopted OmniDocBench (v1.0) benchmark and present results in Table~\ref{omnidocbench}. We report the results on English subset. For the formula metrics, we should note that since Nemotron-Parse outputs text in markdown format, simple mathematical equations not requiring specialized LaTeX commands (e.g., consisting of purely subscripts and superscripts as additional formatting) would be represented by markdown text rather than enclosed in LaTeX math environment delimiters, resulting in their penalization in 'formula' category of OmniDocBench, even in cases of correct representation via markdown. As can be seen, Nemotron-Parse and Nemotron-Parse-TC achieve competitive performance, showing strong overall accuracy, and in particular accuracy on tables and reading order metrics, outperforming competing methods in the same size/vision token count category. We note that thanks to vastly improved reading order of Nemotron-Parse-TC, it outperforms the base Nemotron-Parse on this benchmark overall, while having only minor losses in other sub-categories.

\begin{table}[]
\small
\begin{tabular}{lcccccc}
\hline
\multicolumn{1}{c}{\multirow{2}{*}{\textbf{Model}}} & \multirow{2}{*}{\textbf{Tokens}} & \multicolumn{5}{c}{\textbf{English}}                                                   \\ \cline{3-7} 
\multicolumn{1}{c}{}                                &                                  & \textbf{overall} & \textbf{text}  & \textbf{formula} & \textbf{table} & \textbf{order} \\
\midrule
\multicolumn{7}{c}{\textbf{Pipeline Models}}  \\ \hline
Dolphin \citep{feng2025dolphin}                                    & -                                & 0.356            & 0.352          & 0.465            & 0.258          & 0.35           \\
Marker \citep{marker}                               & -                                & 0.296            & 0.085          & 0.374            & 0.609          & 0.116          \\
Mathpix \citep{mathpix}                                 & -                                & 0.191            & 0.105          & 0.306            & 0.243          & 0.108          \\
MinerU-2.1.1 \citep{wang2409mineru}                               & -                                & 0.162            & 0.072          & 0.313            & 0.166          & 0.097          \\
MonkeyOCR-1.2B  \citep{li2025monkeyocr}                          & -                                & 0.154            & 0.062          & 0.295            & 0.164          & 0.094          \\
PPstructure-v3  \citep{cui2025paddleocr}                          & -                                & 0.152            & 0.073          & 0.295            & 0.162          & 0.077          \\ 
\midrule
\multicolumn{7}{c}{\textbf{End-to-end Models}     } \\ \hline
Nougat    \citep{Nougat}                               & 2352                             & 0.452            & 0.365          & 0.488            & 0.572          & 0.382          \\
SmolDocling   \citep{nassar2025smoldocling}                 & 392                              & 0.493            & 0.262          & 0.753            & 0.729          & 0.227          \\
InternVL2-76B  \citep{chen2024far}                             & 6790                             & 0.44             & 0.353          & 0.543            & 0.547          & 0.317          \\
Qwen2.5-VL-7B   \citep{bai2025qwen2}                            & 3949                             & 0.316            & 0.151          & 0.376            & 0.598          & 0.138          \\
OLMOCR       \citep{poznanski2025olmocr}                              & 3949                             & 0.326            & 0.097          & 0.455            & 0.608          & 0.145          \\
GOT-OCR2.0    \citep{GOT}                             & 256                              & 0.287            & 0.189          & 0.360            & 0.459          & 0.141          \\
OCRFlux-3B \citep{ocrflux}                                & 3949                             & 0.238            & 0.112          & 0.447            & 0.269          & 0.126          \\
GPT4o  \citep{openai2023gpt}                                  & -                                & 0.233            & 0.144          & 0.425            & 0.234          & 0.128          \\
InternVL3-78B  \citep{zhu2025internvl3}                          & 6790                             & 0.218            & 0.117          & 0.38             & 0.279          & 0.095          \\
Qwen2.5-VL-72B    \citep{bai2025qwen2}                          & 3949                             & 0.214            & 0.092          & 0.315            & 0.341          & 0.106          \\
dots.ocr   \citep{dotsocr}                               & 3949                             & 0.182            & 0.137          & 0.320            & 0.166          & 0.182          \\
Gemini2.5-Pro    \citep{gemini}                           & -                                & 0.148            & 0.055          & 0.356            & 0.13           & 0.049          \\
MinerU2.0   \citep{wang2409mineru}                               & 6790                             & 0.133            & 0.045          & 0.273            & 0.15           & 0.066          \\
dots.ocr$^{\dagger\text{200dpi}}$ \citep{dotsocr}         & 5545                             & 0.125            & \textbf{0.032} & 0.329            & \textbf{0.099} & \textbf{0.04}  \\
DeepSeek-OCR-Tiny    \citep{wei2025deepseek}                                            & 64                               & 0.386            & 0.373          & 0.469            & 0.422          & 0.283          \\
DeepSeek-OCR-Small     \citep{wei2025deepseek}                                           & 100                              & 0.221            & 0.142          & 0.373            & 0.242          & 0.125          \\
DeepSeek-OCR-Base     \citep{wei2025deepseek}                                            & 256(182)                         & 0.137            & 0.054          & 0.267            & 0.163          & 0.064          \\
DeepSeek-OCR-Large  \citep{wei2025deepseek}                                              & 400(285)                         & 0.138            & 0.054          & 0.277            & 0.152          & 0.067          \\
DeepSeek-OCR-Gundam  \citep{wei2025deepseek}                                             & 795                              & 0.127            & 0.043          & 0.269            & 0.134          & 0.062          \\
DeepSeek-OCR-Gundam-M$^{\dagger\text{200dpi}}$ \citep{wei2025deepseek}                   & 1853                             & \textbf{0.123}   & 0.049          & \textbf{0.242}   & 0.147          & 0.056          \\ \midrule
\textbf{Nemotron-Parse }                                             & 3201                             & 0.131            & 0.052          & 0.288            & 0.118          & 0.066          \\ 
\textbf{Nemotron-Parse-TC }                                             & 833                             & 0.129            & 0.055          & 0.295            & 0.121          & 0.048          \\ \bottomrule
\end{tabular}
\caption{Accuracy of Nemotron-Parse on OmniDocBench. Metrics for competing methods are obtained from DeepSeek-OCR paper \citep{wei2025deepseek}.}\label{omnidocbench}
\end{table}

\subsection{Table extraction}

Nemotron-Parse and Nemotron-Parse-TC exhibit strong table performance compared to similarly-sized and even larger models, as seen from the OmniDocBench results. We additionally report TEDS and S-TEDS on several public table benchmarks in Table~\ref{tedssteds}. Since Nemotron-Parse predicts tables in LaTeX format, we convert the predicted tables to HTML or Markdown where necessary.

Further, we compare Nemotron-Parse to other competing methods on public RD-TableBench \citep{rdbench} benchmark which consists of a collection of a diverse in-the-wild tables. We report the results in Table~\ref{rdbench}. As can be seen, Nemotron-Parse achieves competitive performance, being outperformed only by Reducto.

\begin{table}[h]
\centering
\begin{tabular}{l|cc|cc}
\toprule
& \multicolumn{2}{c|}{\textbf{Nemotron-Parse}} & \multicolumn{2}{c}{\textbf{Nemotron-Parse-TC}} \\
& TEDS & S-TEDS & TEDS & S-TEDS\\
\midrule
RD-TableBench & 86.2 & 79.9 & 85.3 & 79.6 \\
PubTabNet &  81.3 & 93.99 & 80.9 & 93.6 \\
OmniDocBench 1.0 en & 82.68 & 89.06 & 84.73 & 91.44 \\
\bottomrule
\end{tabular}
\caption{TEDS and S-TEDS of Nemotron-Parse on public table extraction benchmarks.}\label{tedssteds}
\end{table}

\begin{table}[h]
\centering
\begin{tabular}{l|ccc}
\toprule
\textbf{Method} & \textbf{Table similarity} \\
\midrule
Reducto & \textbf{90.2} \\
Azure & 82.7 \\
Textract & 80.9 \\
Sonnet 3.5 & 80.7 \\
GPT-4o & 76.0 \\
Llamaparse & 74.6 \\
Gcloud & 64.6 \\
Unstructured & 60.2 \\
\midrule
\textbf{Nemotron-Parse} & 85.8 \\
\textbf{Nemotron-Parse-TC} & 85.4 \\
\bottomrule
\end{tabular}
\caption{Table extraction accuracy on RD-TableBench benchmark. Metrics for competing methods are obtained from public Reducto results. \citep{rdbench}}\label{rdbench}
\end{table}

\subsection{Multilingual OCR}

Nemotron-Parse supports a variety of European languages, as well as other languages in limited domains (due to the nature of the data sources in the training blend for Chinese, Japanese, and Korean we find Nemotron-Parse to perform well in the scientific domain as well as standard pdf documents, with limited support for in-the-wild images/documents in these languages). To assess the multilingual capability of Nemotron-Parse, we evaluate it on a set of multilingual documents obtained via NVpdftex pipeline \citep{NVpdftex-pipeline}. Test set of each language consists of 10,000 dense documents from scientific domain, with font and color augmentations ensuring the document diversity, and we report plaintext WER and F1 in Table~\ref{multiling}. As can be seen, Nemotron-Parse achieves competitive F1 > 0.96 on all languages, with accuracy on English achieving 0.98. % Our model offers strong support for English and many European languages.

\begin{table}[h]
\centering
\begin{tabular}{l|cc}
\toprule
\textbf{Language} & \textbf{WER} ↓& \textbf{F1} ↑ \\
\midrule
English & 0.03 & 0.98 \\
German & 0.06 & 0.96 \\
French & 0.05 & 0.97 \\
Italian & 0.05 & 0.97\\
Spanish & 0.04 & 0.97 \\
Chinese & 0.03 & 0.98 \\
Japanese & 0.03 & 0.98 \\
\bottomrule
\end{tabular}
\caption{OCR results on multilingual NVpdftex dataset.}\label{multiling}
\end{table}

% \subsection{Multi-token inference}
% \begin{table}[h]
% \centering
% \setlength{\tabcolsep}{5pt}
% \begin{tabular}{l|c|cccc}
% \toprule
%                   {Method}                      &       {$\frac{tkn}{step}$}            & {WER} $\downarrow$  & {F1} $\uparrow$ \\ %& $\frac{sec}{img}$ $\downarrow$ & $\frac{sec}{100}$ $\downarrow$  \\ \midrule
%                   % Nougat \citep{Nougat} & 1 & - & - & 4.7 & 0.41 \\ 
%                   % GOT \citep{GOT} & 1 & 0.25 & 0.82 & 9.8 & 0.90 \\ 
% \multirow{2}{*}{Nemotron-Parse-2tkn \hl{todo: speed}} & 2 & 0.13          & 0.95  \\ %&  &  \\
%                                              & 1 & \textbf{0.12} & \textbf{0.95} \\ %&  & \\
%                                          \bottomrule
% \end{tabular}
% \vspace{-2mm}
% \caption{Results and speed of multi-token models and competing methods. We report the average speed per image on \benchds test set ($\frac{sec}{img}$), and speed per 100 tokens ($\frac{sec}{100}$). These values are obtained from a PyTorch-based inference pipeline on an NVIDIA H-100 GPU. \vspace{-15pt}}\label{tab:multi-token-results}
% \end{table}

\section{Implementation}
We release both Nemotron-Parse-v1.1 and Nemotron-Parse-v1.1-TC on Huggingface in fp32/bf16 formats at \url{https://huggingface.co/nvidia/NVIDIA-Nemotron-Parse-v1.1} and \url{https://huggingface.co/nvidia/NVIDIA-Nemotron-Parse-v1.1-TC} along with VLLM support. Nemotron-Parse is additionally released as an optimized NIM container at \url{https://build.nvidia.com/nvidia/nemotron-parse}. 

We report the speed on a single H100 GPU of bf16 model in the Table~\ref{tab:speed}. Inference speed is measured as average tokens/seconds end-to-end over 10,000 pages of length 1000 tokens each. This roughly translates to 5 pages/second for Nemotron-Parse-TC and 4 pages/second for Nemotron-Parse for an average page.
\begin{table}[h!]
\centering
\begin{tabular}{c|c}
\hline
\textbf{Method} & \textbf{Tok/sec}  \\ \hline
Nemotron-Parse & 3800  \\ \hline
Nemotron-Parse-TC & 4500  \\ \hline
\end{tabular}
\caption{ Inference throughput (tokens per second) showing that Nemotron-Parse-TC achieves higher generation speed than the standard Nemotron-Parse mode}\label{tab:speed}
\end{table}

\section{Contributors}

\textbf{Core Model Development }
Kateryna Chumachenko, Amala Sanjay Deshmukh, Jarno Seppänen, Ilia Karmanov, Chia-Chih Chen, Lukas Vögtle, Philipp Fischer, Marek Wawrzos, Timo Roman, Karan Sapra, Andrew Tao, Bryan Catanzaro

\textbf{Infrastructure, Data, Benchmarking, and Product}

Saeid Motiian, Roman Ageev, Kedi Wu, Alexandre Milesi, Maryam Moosaei, Krzysztof Pawelec, Padmavathy Subramanian, Mehrzad Samadi, Xin Yu, Celina Dear, Sarah Stoddard, Jenna Diamond, Jesse Oliver, Leanna Chraghchian, Patrick Skelly, Tom Balough, Yao Xu, Jane Polak Scowcroft, Daniel Korzekwa, Darragh Hanley, Sandip Bhaskar

\section{Acknowledgement}
 
 We would like to thank Mike Ranzinger, Collin McCarthy, Bo Liu, Jean-Francois Puget, Sean Sodha, Nave Algarici, Randy Gelhausen for helpful discussion and feedback.

\section{Examples}

In the following section, we provide several output examples showcasing the capabilities of Nemotron Parse v1.1 in layout understanding, table extraction, OCR, and formula extraction.

\begin{figure}[h!]
  \centering
  \includegraphics[width=0.72\textwidth]{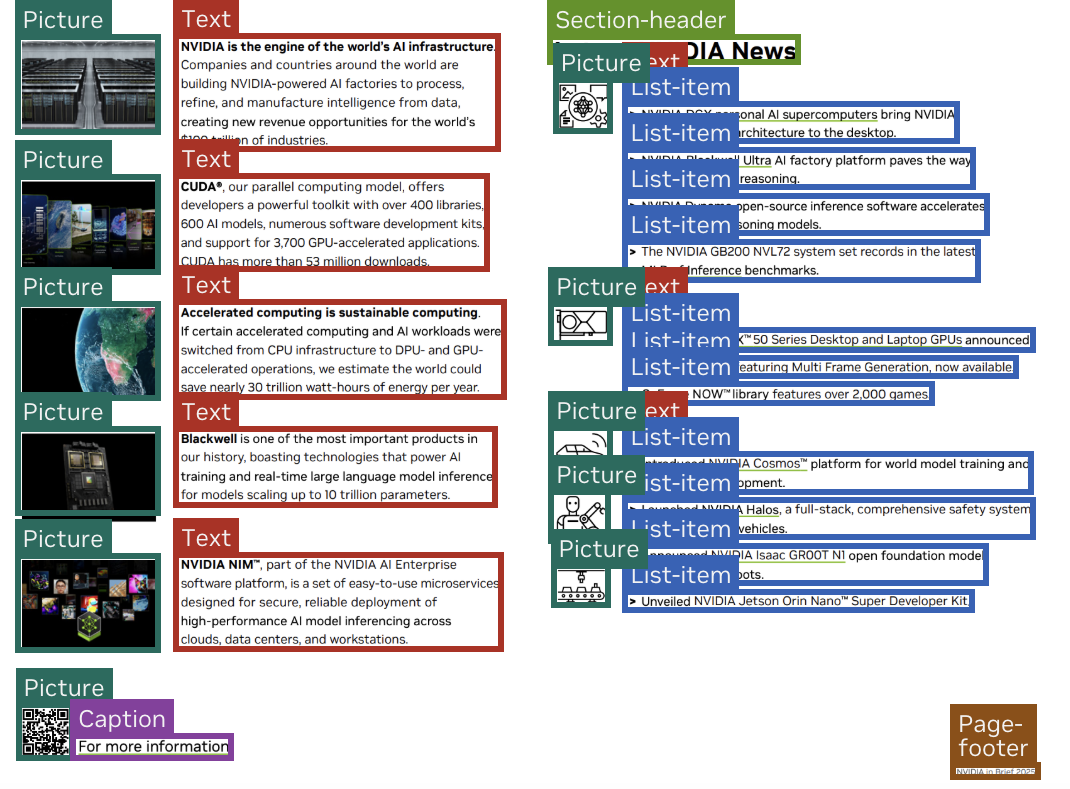}
  \caption{Layout analysis: bounding box detection and prediction of semantic classes}
\end{figure}

\begin{figure}[h!]
  \centering
  \includegraphics[width=\textwidth]{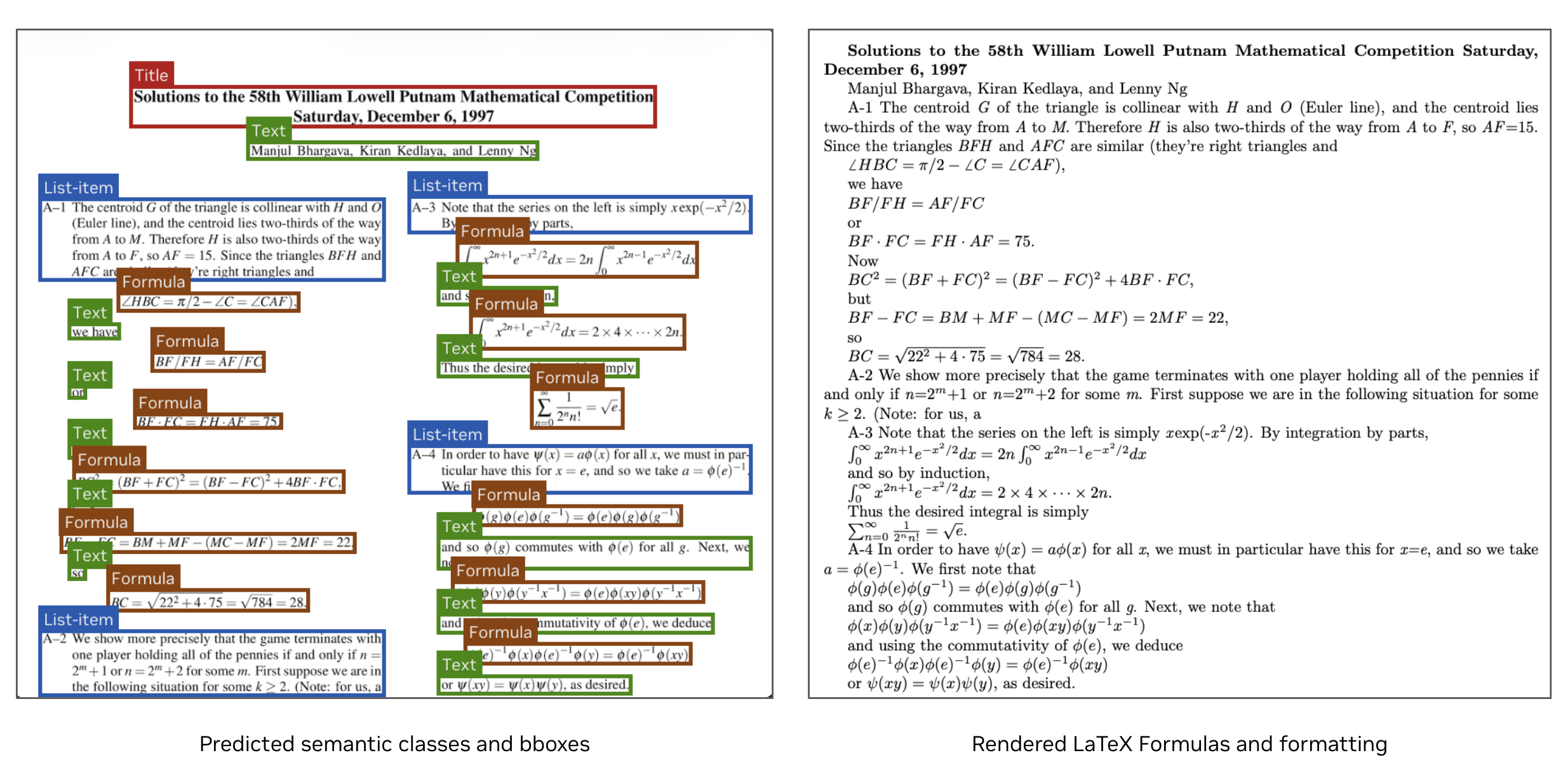}
  \caption{OCR, extraction of text formatting and mathematical equations in LaTeX and markdown.}
\end{figure}

\begin{figure}[h!]
  \centering
  \includegraphics[width=\textwidth]{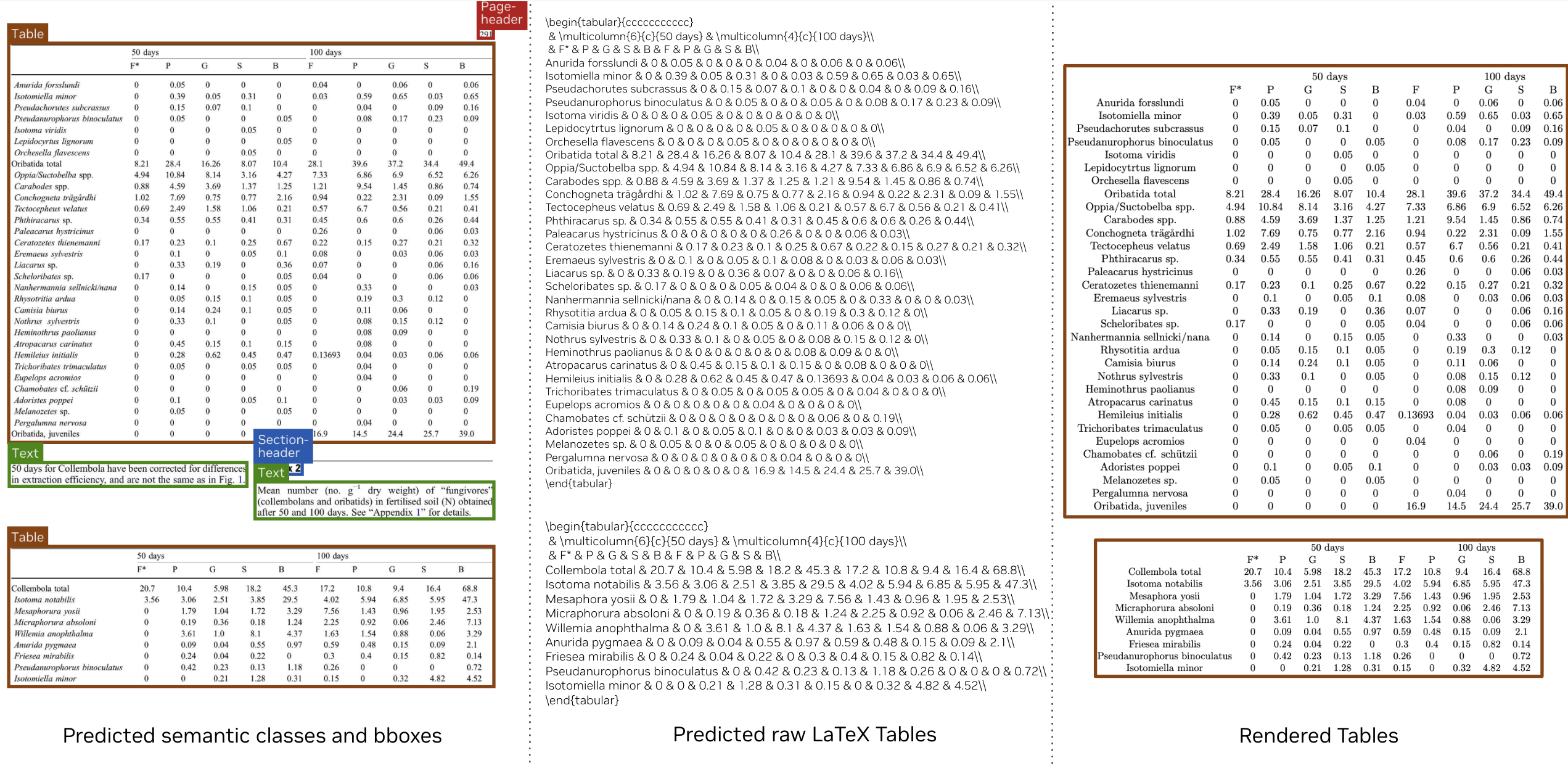}
  \caption{Extraction of complex tables to LaTeX format.}
\end{figure}

% \begin{figure}[h!]
%   \caption{}
%   \centering
%   \includegraphics[width=0.5\textwidth]{assets/ex2.png}
%   \caption{Layout analysis: bounding box detection and prediction of semantic classes}
% \end{figure}

\bibliography{references}
\bibliographystyle{references}

\end{document}